# Efficient sign language recognition system and dataset creation method based on deep learning and image processing

A. L. Cavalcante Carneiro[a], L. Brito Silva[a], D. H. Pinheiro Salvadeo[a]

[a]Dept. of statistics, applied mathematics, and computation, State Univ. of São Paulo, Av. 24 A, 1515, Rio Claro, SP, Brazil, 13506-700

**ABSTRACT**

New deep-learning architectures are created every year, achieving state-of-the-art results in image recognition and leading to the belief that, in a few years, complex tasks such as sign language translation will be considerably easier, serving as a communication tool for the hearing-impaired community. On the other hand, these algorithms still need a lot of data to be trained and the dataset creation process is expensive, time-consuming, and slow. Thereby, this work aims to investigate techniques of digital image processing and machine learning that can be used to create a sign language dataset effectively. We argue about data acquisition, such as the frames per second rate to capture or subsample the videos, the background type, preprocessing, and data augmentation, using convolutional neural networks and object detection to create an image classifier and comparing the results based on statistical tests. Different datasets were created to test the hypotheses, containing 14 words used daily and recorded by different smartphones in the RGB color system. We achieved an accuracy of 96.38% on the test set and 81.36% on the validation set containing more challenging conditions, showing that 30 FPS is the best frame rate subsample to train the classifier, geometric transformations work better than intensity transformations, and artificial background creation is not effective to model generalization. These trade-offs should be considered in future work as a cost-benefit guideline between computational cost and accuracy gain when creating a dataset and training a sign recognition model.

## 1. INTRODUCTION

Communication is fundamental in any society and culture, which people daily use to express themselves and have access to the most basic services like public transport, school, and health care. Sign language is the form of communication used in all countries by people with severe hearing loss, a condition that reaches, at some level, around 466 million people worldwide[19]. The problem is that most hearing people do not know how to speak by signs, creating a barrier that makes it difficult for the deafs to have social interactions.

The areas of computer vision and deep learning are advancing rapidly through the creation of new algorithms, which can be used as a novel interface for human communication, ensuring accessibility for the hearing impaired. Although there are different sign languages around the world, the same deep-learning algorithm can be generalized to all of them with just minimal training and hyperparameters changes. Nevertheless, the dataset creation must be done for all languages, which is complicated, expensive, and time-consuming, being desirable to find techniques that contribute to optimize this process.

Most of the existing work in sign recognition focuses only on the hands, which limits detection to alphanumeric signs or simple hands configurations, compelling deafs to use the spelled speech, which is difficult to understand and impractical. However, to perform word detection it is necessary to consider the entire image, once the meaning of a word depends not only on the hand configuration but also on its position concerning other parts of the body.

Based on this, our work aims to create a Convolutional Neural Network (CNN) algorithm to classify different words in Brazilian Sign Language (BSL) and to investigate the techniques of digital image processing and machine learning that more contributes to accuracy gain, considering a reasonable computational cost and a cheap dataset, created using a smartphone camera in RGB color system, few interpreters and the same background.

## 2. RELATED WORK

We can categorize previous works in this area by the different sensors used to capture the signs: data gloves, depth sensors, and photographic sensors. Data gloves are hardware devices[14] attached to the user in a glove format to capture the hand movements. The problem with these gadgets is they are expensive, not practical to use daily, and must be unique for each user given the differences in hand morphology. Furthermore, these gloves only capture hand information, limiting the ability to recognize words.

Another approach is the use of depth sensors like leap motion or Microsoft Kinect, but just the second one is adequate to word recognition, once it is not only limited to hands. Machado[10] created a dataset containing over 21,000 images extracted from RGB-D videos captured using Kinect and achieved 79.80% accuracy with a 3-D CNN to recognize words in BSL. However, this is still an expensive and not practical sensor and requires more computational power due to the extra depth channel.

The last approach is based on photographic sensors like smartphone cameras. The use of these sensors makes data acquisition harder[12] but with the advantage of being less costly, very common, and practical to use in the quotidian. Magalhães[11] created a BSL dataset with over 660 videos using a mobile camera, and achieved 99,8% accuracy using data augmentation techniques and creating its own CNN architecture, but the details concerning the dataset creation were not explored. Rao *et al.*[15] created a dataset containing over 200 signs in Indian sign language, but no data augmentation was used and the dataset was limited to signs performed with one hand. Ji *et al.*[8] classified signs performing image sampling and using grayscale, achieving an accuracy of 86%, however just 6 signs were recognized, and no further exploration was done concerning data augmentation.

Huang *et al.*[7] used a 3-D CNN framework and RGB videos to form sentences using Hierarchical Attention Network (HAN) and Latent Space for temporal information and word alignment, but they do not use image transforming to augment data, which could improve the 82.7% accuracy reached by the authors. Finally, Cruz[2] used a variant of 3-D CNN with transfer learning and data augmentation techniques to reach 83.75% in a large BSL sign dataset with more than 23,000 images and 560 different signs. The transformations applied were a good starting point, but a better exploration could be made, such as changing the background (once the background of all the dataset videos was the same), and the test of the most impactful data augmentation techniques for sign recognition. In general, previous works do not explore details about the data acquisition, augmentation, and model generalization, which can help future research to create large-scale datasets and sign translation systems.

## 3. MATERIAL AND METHODS

### 3.1 Dataset creation

Our dataset was created using different smartphone cameras in the RGB color system, facilitating the reproduction of the method once the sensor is accessible. We captured 14 different signs that are used daily by deaf people, shown in table 1, where each sign was repeated 3 times by 2 different interpreters due to small variations present in each sign execution, totaling 84 videos.

Table 1. Words that compound dataset.

| Category | Words |
|---|---|
| Pronoun | I, my, you, we, you (plural), your. |
| Noun | woman, man, car, motorcycle, bus, supermarket, hospital, bank. |

We took the videos during the day using natural light and taking care of the brightness and light reflections. The captured resolution was 1920x1080 pixels, but the frames were later rescaled to 331x331 pixels to reduce the computational cost, and the background was the same for all the videos, which can be a disadvantage to the model generalization in a real-world scenario[2], but it is considerably faster and easier than changing the scene programmatically. Therefore, artificial background creation may be an important technique to leave faster dataset acquisition while decreasing model bias.

We initially recorded at 60 Frames Per Second (FPS) to test the performance of the model using different frame rates by subsampling the original video, which can lead to distinct results[2]. Capturing videos at 60 FPS requires a better sensor and more storage space, but can mitigate the motion blur during the sign execution, being an important trade-off to consider. Based on this, we created another dataset following the same steps as the first one, but capturing the signs at 30 FPS, to test if there is a significant difference between the FPS used to record the sign.

Since we will use a 2-D CNN architecture it is necessary to get the frames individually instead of using the entire video, thus, an algorithm was developed to subsample the frames in 60 to 30 and 20 FPS totaling 18300, 9299, and 6244 images respectively. In addition to all the words in table 1, a new category was created, called "formation" which are signs that have no meaning because the hand is beginning or finishing the sign formation, or the interpreter is in the rest position, waiting to do the sign, as illustrated in figure 1.

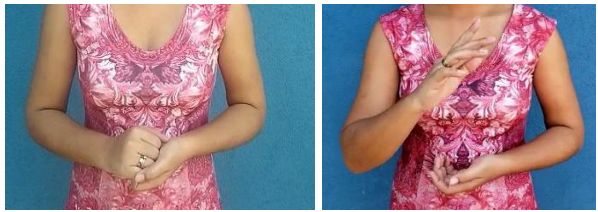

Figure 1. Rest position and sign formation respectively, both categorized as "formation" in our dataset.

Lastly, a validation set was created repeating the same 14 signs twice by 2 interpreters (56 videos and 3782 images at 60 FPS) but in different backgrounds and light conditions, to test the performance and generalization of the model in more adverse situations.

**3.2 Data augmentation**

Data augmentation is a technique used to modify the dataset samples using transformations, which is important to reduce overfitting and make the model more robust[5], improving its performance[2,3]. In this context, it is especially useful to work with different data augmentation techniques, to make it cheaper to create a sign language dataset. We apply geometric and intensity transformations and investigate the impact of each transformation on the model accuracy, as some operations may not work properly depending on the problem and the dataset[3], introducing classification errors, and every new transformation includes extra training time[2], requiring the choice of appropriate techniques.

Geometric transformations are characterized by mapping the pixel position in the original image to a new position in the resulting image. Rotation, translation, shear, and scale were the techniques used for these transformations[6]. On the other hand, intensity transformations are modifications in the pixel value by applying mathematical formulas and changing the brightness and color conditions[6]. The transformations and their possible range of application are illustrated in table 2 and were done randomly in training time.

Table 2. Transformations that were applied to augment data and the possible ranges of application.

| Transformation type | Range |
|---|---|
| Zoom | -0.2 - 0.2% |
| Rotation | 0 - 30º |
| Shear | 0 - 10° |
| Translation | True |
| Brightness | 0.5% - 1.2% |
| Height Shift | 0 - 0.15º |
| Channel Inversion | 0.8 - 1.2% |

The application ranges were chosen empirically based on tests and assumptions about the data. For example, the ranges of zoom and height shift transformations can not be greater than those chosen for our dataset (where all the interpreters are centered in the image), as this would cause the hands to be cut. Also, it does not make sense to use over 30 degrees in the rotation range, once it can deform the signal, creating a new one and because we assume the interpreters were in the vertical position.

Background replacement can also be considered as a data augmentation technique but it was not done in training time due to computational cost. To perform this task, semantic segmentation was used with transfer learning from the state-of-the-art system DeepLabV3[1]. The pixels corresponding to the segmented person are preserved but the background is replaced by a new one, randomly chosen among five possible scenes.

### 3.4 Model training

The dataset was split into 80% of samples for training and 20% for testing, once it is a common dataset partition ratio[5] and accuracy was the evaluation metric, as observed in other works[2, 10, 11, 15], given by the division between the number of correct predictions and the total number of samples. Transfer learning is another important technique used to improve model performance[2] and decrease the training time, even with a small dataset[5]. We transferred the learned parameters from XCeption[4] architecture trained on ImageNet[9] to our network, replacing only the last layers to address the given problem.

The training was carried out for 25 epochs with a batch size of 128, which showed empirically to be enough to train the model, using a P100 GPU as dedicated hardware. Every test was repeated 3 times in the same setup, capturing the average results and comparing them statistically, using analysis of variance (ANOVA) to compare 3 or more results and T-student, when the comparison is between two groups.

As a final test for the findings of this paper, we created and fine-tuned a multi-stream CNN architecture, having one channel to extract the features of the entire image and another two channels for each hand, forcing the model to focus on local and global information. Each stream of the CNN was based on EfficientNet-B0[17] with an image resolution of 224x224 pixels, to keep the number of parameters low. We also trained an object detection algorithm using a public dataset[13] and based on EfficientDet[18], to separate the hand of the image and pass it to the feature extractor, which is later merged into a multi-layer perceptron, as figure 2 illustrates.

We initially trained the multi-stream CNN for 20 epochs with a learning rate of 10e-3. To fine-tune, we unfreeze about 60% of each CNN channel, training again for 20 epochs but with a learning rate of 10e-6.

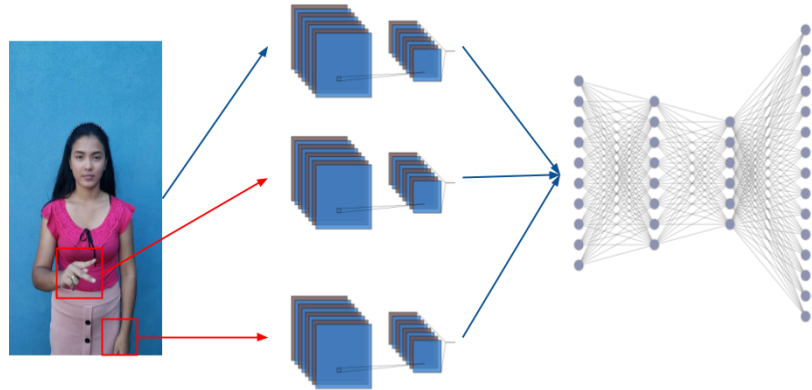

Figure 2. The multi-stream CNN architecture created to capture local and global information in the image.

## 4. RESULTS AND DISCUSSION

**4.1 Investigating data augmentation techniques**

We trained the model with each data augmentation technique individually, shown in table 2, to search for the more suitable options for the sign recognition task. Table 3 shows the average accuracy of the operations.

Table 3. Average accuracy comparison between each data augmentation technique trained individually, considering the 20 FPS dataset.

| Transformation type | Test set | Validation set |
|---|---|---|
| Zoom | 90.65 | **56.93** |
| Rotation | 87.23 | **61.22** |
| Shear | 90.07 | **53.31** |
| Translation | 89.87 | 49.49 |
| Brightness | 91.20 | 49.91 |
| Height Shift | 89.90 | **55.63** |
| Channel Inversion | 91,17 | 50,94 |

As the main goal of data augmentation is to improve the model invariance, we focus our attention on the results of the validation set, where there is a statistical difference, proved by the ANOVA test with a p-value of 0.00039. Thus, we conclude that in general, geometrical transformations perform better than intensity transformations, which make sense considering how a human sees the sign, where the shape and spatial location has real importance in the word recognition, different from color, background, and clothes.

With these results, in table 4, we compare the model subjecting it to various circumstances.

Table 4. We trained the model without and with data augmentation as baselines. Then, we trained with the best performance techniques (highlighted in table 3) and with the worst performance (not highlighted in table 3), to compare the results.

|  | Test set | Validation set |
|---|---|---|
| Without augmentation | 91.46 | 52.37 |
| All augmentation techniques | 85.50 | 60.32 |
| Best performance techniques | 85.50 | 60.97 |
| Worse performance techniques | 90.01 | 51.29 |

Table 4 highlights that just using the data augmentations with the best individual performance is enough to achieve similar results as using all the techniques, maintaining the pattern observed individually. On the other hand, the data augmentation techniques with a worse individual performance prove statistically (with a p-value of 0.0396 in the T-student test) to degrade the model performance.

Although the accuracy decreased in the test set when using data augmentation, by the invariance introduced in the model, the effectiveness of the method is proven due to the increased performance of the validation set and also by the reduction of overfitting as shown in figure 3.

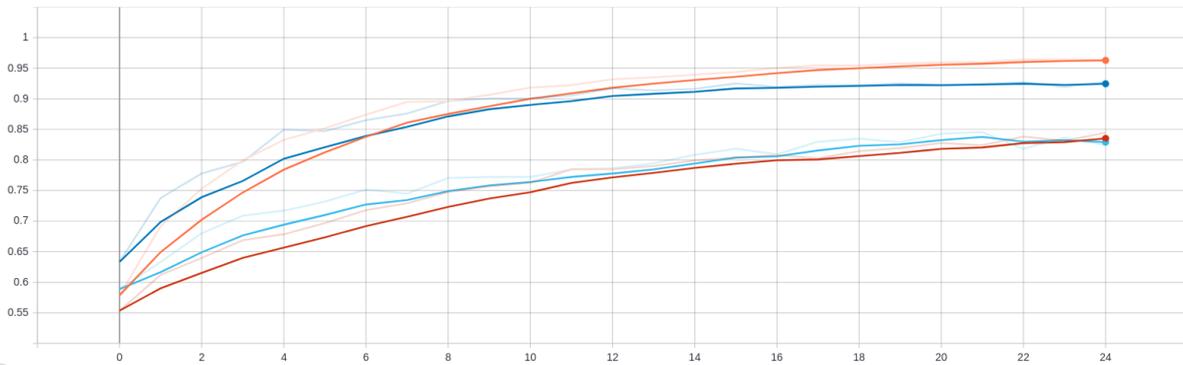

Figure 3. Accuracy without and with data augmentation in the 20 FPS, showing how it mitigates train (orange) and test (blue) variance.

**4.2 Frames per second comparison**

At first, we studied the most suitable FPS to subsample the videos into images to train the model by recording at 60 FPS. Table 5 brings the results.

Table 5. Average model accuracy over different subsamples from the videos captured at 60 FPS.

| FPS Rate | Test set | Validation set |
|---|---|---|
| 60 | 91.14 | 50.27 |
| 30 | 90.09 | 59.65 |
| 20 | 85.50 | 60.97 |

Remarkably, 60 FPS does not compensate for the computational resource required, since it obtained about 10% less accuracy in the validation set. This is probably because this frame rate has almost 2 and 3 times more images than 30 and 20 FPS, as figure 2 showed, which may contribute to overfitting, causing a greater variance in the validation set. Besides that, the consecutive images of a video are similar to each other, generating a low gain of information.

The T-student test reveals that 30 and 20 FPS have a significant difference in the test set, leading to the conclusion that it is the best choice for this situation, but this should vary depending on the dataset size, once the training time is bigger for 30 FPS, and the exploration of spatio-temporal features, influencing the amount of information needed to be extracted from the video.

The last test involving frame rates was to compare the performance between a dataset captured at 30 FPS and another one captured at 60 FPS, shown in table 6.

Table 6. Average accuracy between datasets captured at 30 and 60 FPS.

|  | Test set | Validation set |
|---|---|---|
| 30 FPS captured at 60 | 90.09 | 59.65 |
| 30 FPS captured at 30 | 85.54 | 60.47 |
| 20 FPS captured at 60 | 85.50 | 60.97 |
| 20 FPS captured at 30 | 76.84 | 55.76 |

In the test set, the results are favorable to the dataset captured at 60 FPS, as the greater number of images helps the model to fit better during training (as shown in table 5). On the other side, In the validation set, there is no significant difference between the dataset capture at 60 or 30 FPS (with a p-value of 0.58). Another relevant fact is that the dataset captured at 30 FPS had fewer images than the captured at 60 FPS and subsampled to 30, owing to the faster execution of the signs, which is a normal variation depending on the interpreter and the situation. Therefore, instructing the interpreters to execute the sign slowly should help to further mitigate these accuracy differences in capture rate, mainly in a well-lit scenery, where the motion blur is less perceptive.

Thus, in an uncontrolled scenario, with different lighting conditions and sign speed execution by the interpreters, capturing the video at 60 FPS and resampling it to 30 FPS should be the best choice, getting a large number of images while avoiding motion blur, with the drawback to require a better sensor and more storage space. Despite that, in a well-controlled scenario, capture at 30 FPS will produce satisfactory results.

**4.3 Artificial background effect**

The background changing was not applied to the 60 FPS set, once we show that this frame rate is not as effective as 20 or 30 FPS. Figure 4 shows the results of the semantic segmentation.

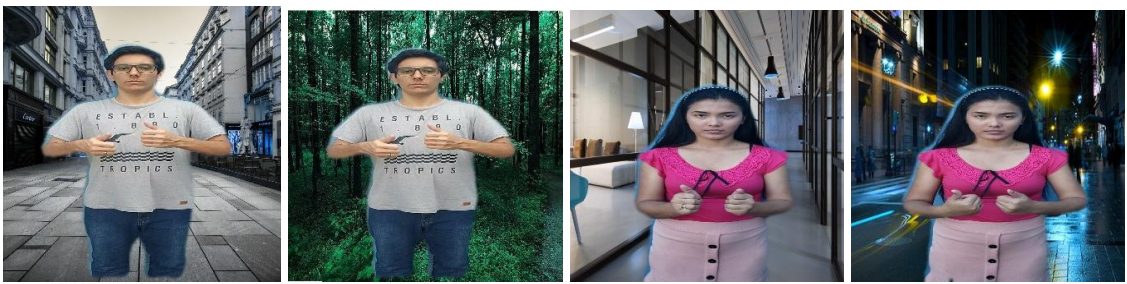

Figure 4. Signs with different backgrounds using the DeepLabV3 segmentation method.

The training was done as before and the results, shown in table 7, highlights that the extra effort to replace the image background is not compensated, due to accuracy worsening.

Table 7. Comparing average model accuracy with artificial background replacement ('background suffix') and without it.

|  | Test set | Validation set |
|---|---|---|
| 30 FPS background | 85.54 | 60.47 |
| 30 FPS | 90.09 | 59.65 |
| 20 FPS background | 76.84 | 55.76 |
| 20 FPS | 85.50 | 60.97 |

To understand more about these results, we used a tool to explain the model predictions called LIME[16], highlighting the parts of the image that contributed more to the inference, as illustrated in figure 5. Explanations suggest that the model is focusing on the correct part of the image, considering the position of the hand of the interpreters to infer the signal. This shows that the background is not biasing the results and that is why the replacement does not aggregate relevant features, acting just like a color transformation, which has a low contribution to the accuracy, as shown in table 7.

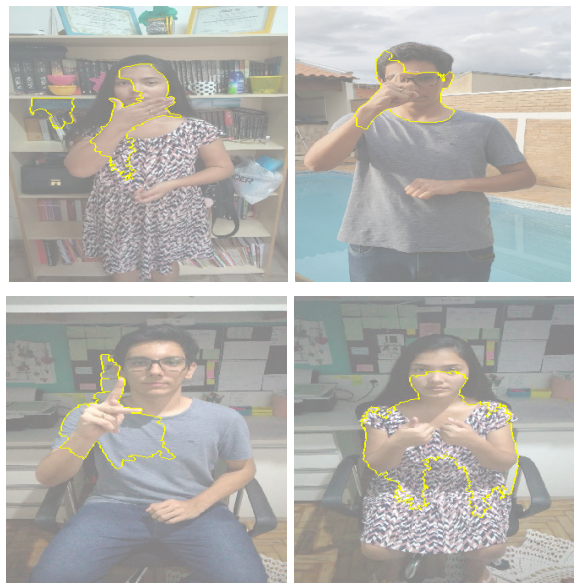

Figure 5. Using LIME to explain model predictions in the validation set.

**4.4 Improving validation accuracy with multi-stream CNN**

Figure 6 shows a confusion matrix of the model trained in the 30 FPS dataset, helping to investigate the mistakes in sign prediction, once the validation accuracy suggests a poor model generalization.

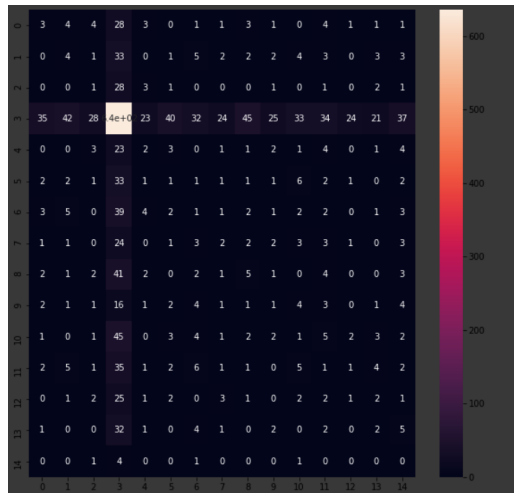

Figure 6. Confusion matrix of the model prediction over the 15 signs.

As observed, most of the errors occur in sign 3, which is the "formation" class, because the transition between this sign and the word is gradual, creating an intersection that can easily confuse the model. To overcome this limitation, we trained the multi-stream CNN to increase the attention of the model for the hands, achieving an accuracy of 96% on the test set and 81% on the validation set, improving the accuracy and generalization of the model to more challenging conditions.

## 5. CONCLUSIONS

In this paper, digital image processing and machine learning techniques were investigated to assist future works involving sign recognition tasks to create a dataset effectively. We show that geometric transformations are more effective than intensity transformations, since the model infers the sign based on the hand shape and position in the image, also being invariant to the background type. We also argued that 60 FPS was the best rate to record the videos of the dataset, but it depends on the context such as the speed of the sign execution and lighting conditions. Equally, when training the model, 30 FPS was a better subsample choice over 60 or 20 FPS but it should vary depending on how the spatio-temporal features are explored. Finally, we created a multi-stream CNN architecture, based on our findings, that achieved an accuracy of 96% on the test set and 81% on the validation set. In future works, a larger dataset can be created, considering the findings of this paper and using different techniques for classification such as a spatio-temporal classifier and ensemble methods.